\documentclass[runningheads,a4paper]{llncs}

\usepackage{amssymb}
\usepackage{graphicx}
\usepackage{epstopdf}

\usepackage{url}
\urldef{\mailsa}\path|{faalobaidli,pnakov}@qf.org.qa|
\urldef{\mailsb}\path|s.j.cox@uea.ac.uk|
\newcommand{\keywords}[1]{\par\addvspace\baselineskip
\noindent\keywordname\enspace\ignorespaces#1}

\begin{document}
 
\mainmatter  % start of an individual contribution

% first the title is needed
\title{Bi-Text Alignment of Movie Subtitles\\
for Spoken English-Arabic\\
Statistical Machine Translation}
% Doctor-Patient Communication in Qatar

% a short form should be given in case it is too long for the running head
\titlerunning{Bi-Text Alignment of Movie Subtitles}

% the name(s) of the author(s) follow(s) next
%
% NB: Chinese authors should write their first names(s) in front of
% their surnames. This ensures that the names appear correctly in
% the running heads and the author index.
%
\author{Fahad Al­-Obaidli$^{\dagger}$
\and Stephen Cox$^{\ddagger}$ \and Preslav Nakov$^{\dagger}$}

%
%\authorrunning{Lecture Notes in Computer Science: Authors' Instructions}
% (feature abused for this document to repeat the title also on left hand pages)

% the affiliations are given next; don't give your e-mail address
% unless you accept that it will be published
\institute{$^{\dagger}$Qatar Computing Research Institute, HBKU\\
$^{\ddagger}$School of Computing Sciences, University of East Anglia, Norwich, UK\\
\mailsa\\
\mailsb\\
%\mailsc\\
%\url{http://www.springer.com/lncs}
}

%
% NB: a more complex sample for affiliations and the mapping to the
% corresponding authors can be found in the file "llncs.dem"
% (search for the string "\mainmatter" where a contribution starts).
% "llncs.dem" accompanies the document class "llncs.cls".
%

\toctitle{Lecture Notes in Computer Science}
\tocauthor{Authors' Instructions}
\maketitle

\begin{abstract}

 We describe efforts towards getting better resources for English-Arabic machine translation of spoken text. In particular, we look at movie subtitles as a unique, rich resource, as subtitles in one language often get translated into other languages. Movie subtitles are not new as a resource and have been explored in previous research; however, here we create a much larger bi-text (the biggest to date), and we further generate better quality alignment for it.
 Given the subtitles for the same movie in different languages, a key problem is how to align them at the fragment level. Typically, this is done using length-based alignment, but for movie subtitles, there is also time information. Here we exploit this information to develop an original algorithm that outperforms the current best subtitle alignment tool, \verb|subalign|. 
 The evaluation results show that adding our bi-text to the IWSLT training bi-text yields an improvement of over two BLEU points absolute.

 %that can as well be aligned using a language independent method, time overlap, which is proven to work better with subtitles than length based alignment. We apply our implementation of the aligner on more than 29K pairs of freely available movie subtitles in Arabic and English and add our generated parallel corpus to the IWSLT'13 English-Arabic SMT baseline. We report an increase of almost 2 BLUE points. Our script also outperforms the current best subtitle alignment tool, \verb|subalign|, by 0.29 BLUE points.

\end{abstract}

\section{Introduction}

Statistical machine translation (SMT) research is continually improving in an attempt to produce systems to meet the ever-increasing  demand for accessibility to content in foreign languages. SMT requires a substantial amount of parallel bi-text in order to build a translation model and unfortunately, such bi-text resources are very limited. Moreover, automatic bi-text alignment is a challenging task.

Existing parallel corpora are mostly derived from specialized domains such as administrative, technical and legislation documents~\cite{koehn2005europarl}. These documents often only cover few widely spoken languages or languages that are either regionally or culturally related. Interestingly, with the huge demand for movie subtitles, movie subtitles online databases are among the fastest growing sources of multilingual data. Many users provide these subtitles for free online in a variety of languages through download services. Subtitles are made available in plain text with a common format to help with rendering the text segments accordingly.

They are usually created to fit pirated copies of copyright-protected movies shared by organized groups, so their use in research could be deemed to be a positive side effect of the Internet movie-piracy scene~\cite{mangeot2005multilingual}. Moreover, the use of these subtitles enables low-cost alignment of multilingual corpora by utilising one of their features, which is temporal indexing of subtitle segments. This approach has created opportunities to align specific language pairs that are difficult to align using the traditional methods or that are of generally scarce resources.

Because of the inherent nature of movie dialogue, subtitles differ from other parallel resources in several aspects. They are mostly transcriptions of material that is often spontaneous speech, which may contain considerable slang language, idiomatic expressions and also fragmental spoken utterances rather than complete grammatical sentences---such material is commonly summarized instead of being literally transcribed. Since these subtitles are user-generated, the translations are free, incomplete and affected by cultural differences. Rephrasing and compression degrees vary between different languages and depend on subtitling traditions. Subtitles also arbitrarily include some information such as the movie title, subtitle author/translator details and trailers. They may also contain translations of visual information such as sign languages. Certain versions of subtitles are especially compiled for the hearing-impaired to include extra information about other, non-spoken sounds such as background noise and, therefore contain material not related to the speech. Furthermore, subtitles must be short enough to fit the screen in a readable manner and to span for a limited time, which creates variable segmentations between languages.

The subtitle languages available differ from one movie to another. Here we are interested in movie subtitles that exist in both English and Arabic. Arabic is a widely-spoken language with 300 million native speakers (and another 120+ million non-native speakers), and has an official status in 28 countries (third in that respect, after English and French). Yet, the digital presence of Arabic is relatively low, compared to what one should expect given the number of speakers. Still, according to web traffic analytics, search queries for Arabic subtitles and traffic from the Arabic region is relatively very high.

The reminder of this paper is organized as follows.
Section~\ref{sec:related} offers an overview of related work. 
Section~\ref{sec:method} presents our fragment-alignment method.
Section~\ref{sec:eval} describes the experiments and the evaluation results.
Section~\ref{sec:discuss} concludes with general discussion and points to possible directions for future work.

\section{Related Work}
\label{sec:related}

There is a body of literature about building multilingual parallel corpora from movie subtitles ~\cite{xiao2009constructing,lavecchiainria00155787}. Tiedemann has put a lot of efforts in this direction and has made substantial contributions for aligning movie subtitles ~\cite{tiedemann2007building,tiedemann07improvedsentence,tiedemann2008synchronizing}. Initially, he gathered bi-texts for 59 languages in his OpenSubtitles2013 corpus, which was obtained from 308,000 subtitles files of around 18,900 movies downloaded from OpenSubtitles.org, one of the free online databases of movie subtitles.

For alignment, Tiedemann, started with the traditional approach of length-based sentence alignment \cite{gale1993program} using sentence boundaries tagged in an earlier stage. This is based on the idea that sentences are linguistically-driven elements and, thus, it would be more appropriate to process them using linguistic features rather than merely aligning subtitles appearing simultaneously on the screen. Nonetheless, there are many untranslated parts of the movie for reasons discussed earlier. The initial results were unsatisfactory.

In follow-up work, Tiedemann took another approach based on the time overlap of subtitle fragments between different languages and made explicit use of the available time information. He also took into account that one subtitle segment can be matched to multiple segments in the target language. 
%He matches N-N pairs by maximising the overlap between possible types. 
This yielded significant improvements. However, some movies in the evaluation set yielded noticeably low matching ratios, which was due to time offsets between subtitles synchronized to different versions of the movie, which consequently radically affected the score even for a slight shift. 

In order to find the right offset and to synchronize two subtitle files, a reference point is required. Tiedemann used cognates between the source and the target subtitle languages as a reference point to calculate the offset where the resulting matches ratio is below a certain threshold. A problem with this approach is that, depending on the language pair, false hit cognates may be used and may instead affect the performance drastically. This approach may not work for English--Arabic anyway due to the lack of enough cognates (cognates are relatively hard to find due to the different writing scripts used). Another approach is to use language dictionaries, which are expensive to build and are also language-dependent, although they do yield better results than the cognate approach.

Volk and Harder~\cite{volk2007} built their Swedish-Danish corpus by first manually translating Swedish subtitles of English TV programmes with the help of trained translators using specified guidelines and, in the process, adding appropriate time information for the translations. Subsequently, Danish translators used the Swedish subtitle files as a template and added their translations of the Swedish subtitles between the available time codes. This ``commercial'' setup allows them to avoid the complex alignment approaches found in other studies ~\cite{tiedemann07improvedsentence,tiedemann2008synchronizing}. They only matched pairs where Swedish and Danish time-stamps differ by less than 0.6 seconds, in order to ensure high quality. Volk and Harder assumed that they could match most of the subtitles except where the Danish translator has changed the time codes sufficiently to fail the strict overlap condition~\cite{volk2009automatic}. They state that their alignment approach is pragmatic and requires minimal human inspection, even though it is not sensible beyond this setup, using ``genesis'' files (Swedish subtitles) as a reference, as demonstrated earlier by Tiedemann~\cite{tiedemann2007building}, where, in reality, many translations require accurate handling.

It is also worth mentioning the AMARA project\footnote{https://www.amara.org/en/} ~\cite{jansen2014amara}, a unique, open, scalable and flexible online collaborative platforms which takes advantage of the power of crowdsourcing and encourages volunteer translation and editing of subtitles of educational videos. 

The core value of the AMARA platform is demonstrated by its ``faster transcription turnaround while maintaining high levels of user engagement.'' This could be achieved by the ease of use of the platform and the ability to remotely transcribe a video without the necessity to re-upload it on the platform. Translation can also be verified by other users.

Abdelali et al.~\cite{abdelali2014amara} used the AMARA content to generate a parallel corpus out of the translations in order to improve an SMT system for the educational domain. They aligned the multi-lingual text of the educational videos and assessed its quality with several measures in comparison with the IWSLT training set, which is very close to the AMARA domain. It is important to note that 75\% of the subtitle segments have exact time values among both languages in the pair. This is due to the neat editing environment that makes it easier for users to use existing subtitles as a template to generate translations for other languages. This also makes it ideal for the alignment process as less effort is required to handle partial overlap. This is not the usual case in more open communities such as movie subtitle platforms, which are merely databases for independently generated subtitles. They report an improvement of 1.6 BLEU points absolute when adding the AMARA bi-text to the IWSLT training bi-text for Arabic-to-English SMT.

\section{Method}
\label{sec:method}

We look at movie subtitles as a unique source of bi-texts in an attempt to align as many translations of movies as possible in order to improve English to Arabic SMT performance. In comparison to other translation directions, translating from English into Arabic in particular is a rare translation direction and often yields significantly lower results when compared with the opposite direction (Arabic to English). 

First, we collected pairs of English-Arabic subtitles of more than 29,000 movies/TV shows, a collection bigger than any pre-existing bilingual subtitles dataset. We then designed a pipeline of heuristic processing in order to eliminate the inherent noise that comes with the subtitles' source in order to yield good quality alignment. We used the time information provided within the subtitle files to develop a ``time-overlap'' based alignment method. Note that time overlap information is language-independent and it has been proven in previous work to outperform other traditional approaches such as length-based approaches, which rely on sentence boundaries to match translation segments~\cite{tiedemann2007building}. We attempted to maximize the number of aligned sentence pairs whilst still maintaining the quality of our parallel corpora to minimize noise caused by alignment errors. 

After producing our parallel corpora, which vary in size and relative quality, we used them as additional training data for an English-Arabic SMT system. We trained the baseline system on the IWSLT13 training bitext only; we compared this system to the same system but retrained with IWSLT13 data plus our newly-produced bi-text. We further used the Arabic subtitles to train a language model. 

Adding each of our alignment models positively impacted the SMT system. In particular, the best system outperforms the baseline by 1.92 BLEU points absolute. Moreover, our method outperformed the results yielded by the best previously available subtitle alignment tool~\cite{tiedemann2012opus} by 0.2 BLUE points absolute.

We provide more detail about the bi-text production below.

\subsection{Data Gathering}

We identified target subtitles from OpenSubtitles using their daily-generated exports, which include references to their entire database of about 3M subtitles (as of May 20th, 2015) along with some useful information. We looked for Arabic and English subtitle files that are provided in the most common SubRip format for consistency. It is common that multiple subtitle versions exist for the same movie in the same language. In this case, we look for movies with matching movie \emph{release} names to form a pair of subtitles for each movie. Otherwise, the versions are picked randomly. It was apparent later that pairs with matching movie release names yielded a significantly higher alignment ratio than those picked at random.

\subsection{Preprocessing}

We then downloaded the subtitles with the kind help of the administrators of OpenSubtitles. The files were then subject to a set of quality checks such as language, format, encoding identifications, and were further cleansed in order to eliminate unnecessary text such as HTML formatting tags. We also split segments of dialogue between multiple speakers (starting with leading hyphens) into multiple segments and split the original allocated time among them according to their new character length.
We excluded corrupt and misidentified subtitle files from our collection, and we ultimately ended up with 29,000 complete pairs of Arabic-English subtitle files of unique movies/TV show episodes.

\subsection{Bi-text Alignment}

We decided to base our bi-text alignment approach on time information given its efficiency with subtitle documents, as found in previous work \cite{tiedemann2007building}. Each segment of one of the language pairs is compared to the segments of the other and the segment time information is used to discover overlaps. When an overlap is identified, a predefined \emph{overlap ratio} threshold is applied to decide whether to count the segment pair as a valid translation or to exclude it from the final corpus. The ratio is a modified version of the Jaccard index and is defined as follows:

\[ratio = \frac{intersect + 1}{union + 1}\]

\noindent where \emph{intersect} is the time period in which both segments overlap over the movie's runtime, while \emph{union} is the period both segments cover together, i.e., the timespan from the start of the earliest segment to the end of the latest one.
\pagebreak

Here is a pseudocode of our approach:

\begin{verbatim}
1. Loop through documents A and B and look for a minimal overlap
2. If no overlap is found proceed
3. If overlap is found calculate the overlap ratio
4. If the ratio is more than or equal to the threshold then
  4.1 Return matching segment pair
  4.2 Note the index of the matching segment in document B
  4.3 Break from the inner toop to continue to the next segment in
document A, comparing it with the next segment in B from the
last noted index.
5. Else, go to 1.
\end{verbatim}

We define and distinguish between overlaps here as (\emph{i})~partial overlaps where the starting time value of a segments lies within the other segment's (from the other document) time interval and its end value is found beyond that interval, and vice-versa, and (\emph{ii})~complete overlaps where both the start and the end time value of one segment lies within the other segment's time interval and vice-versa.

This framework allows us to decide at which side of the pair to iterate next given the kind of overlap. Furthermore, the overlap is only considered a match between two overlapping segments when a predefined minimal overlap ratio threshold is met. We define this ratio as follows: given the start and the end time values for the English and the Arabic segment, we identify the minimum (soonest) and the maximum (latest) start times, $startMin$ and $startMax$, respectively and regardless of the language. And we do the same for the minimum and for the maximum end times as well, $endMin$ and $endMax$. Then we decide the amount of overlap/intersect between the segment $(endMin-startMax)$ by the total/union time covered by both $(endMax - startMin)$ using the formula above.

The ratio provides an indexation of the magnitude of the overlap given the length of the segments. Setting a threshold below 1 (strictest) allows a window for partial and complete overlap cases, described earlier, with excess trailing or leading time intervals beyond or before the actual overlapping interval, respectively, to be matched. This tolerance window is essential for the nature of our data since subtitles are typically compiled with independently defined time values and, thus, are positioned with organic variations between different versions along the movie runtime. And, as described earlier, it is less likely that two versions will have an exact number of segments with exact timestamps.

Finally, we extend the above algorithm to handle not only 1-1 matches but also 1:M ones. We simply look for 1-1 matches as usual and, in case the segments do not meet the minimum overlap ratio threshold and depending on the state of the false overlap, we will expand one side of the segments in the pair by concatenating its time value with the next segment on its side and, hence, calculate a new overlap ratio. The expansion continues on one side of the pair up to five adjacent segments, 
%i.e., 1-5 segments alignment, 
and then stops if no matches are found, and continues searching for 1-1 matches again and so on. If multiple segments match, they are concatenated into one line and white-space separated. 

We only consider expanding one side and not both in a single match, i.e.,~one-to-many (1-M) and not many-to-many (M-M). This is because, for example, two parallel and adjacent segments might not match separately as 1-1, but, when concatenated into 2-2, the longer time intervals might compensate for a higher overlap ratio and, consequently, falsely yield a match. It is even more likely to harvest false matches when expanding further, i.e., 2-3, 3-3, etc. After all, it is less likely that both documents contain a segmentation of the same sentence. Yet, if so, longer sentences that had to be segmented still have a good chance of being matched by the 1-1 alignment condition separately.

\section{Experiments and Evaluation}
\label{sec:eval}

Below we first describe the baseline system and the evaluation setup; then, we present our subtitle alignment and machine translation experiments.

\subsection{Baseline System}

We built a phrase-based SMT model \cite{Koehn:2003:SPT}, as implemented in the Moses toolkit \cite{koehn2007moses}, to train an SMT system translating from English to Arabic. We trained all components of the system (translation, reordering, and language models)
on the IWSLT'13 data,\footnote{The data can be found at http://workshop2013.iwslt.org} which includes a training bi-text of 150K sentences \cite{IWSLT:2013}.

Following \cite{KholyH12},
we normalized the Arabic training, development and test data using MADA~\cite{MADA},
fixing automatically all wrong instances of \emph{alef}, \emph{ta marbuta} and \emph{alef maqsura}.
We further segmented the Arabic words using the Stanford word segmenter \cite{monroe2014word}.
For English, we converted all words to lowercase.

We built our phrase tables using the standard Moses pipeline with max-phrase-length of 7 and Kneser-Ney smoothing.
We first word-aligned the training bi-text using IBM model 4 \cite{brown93mathematic} on the English-Arabic and on the Arabic-English directions; then, we consolidated the two alignments using the grow-diag-final-and symmetrization heuristics. Then, we trained a phrase-based translation model with the standard features and a maximum length of 7.

We also built a lexicalized reordering model \cite{IWSLT:2005}: \emph{msd-bidirectional-fe}.
We further built a 5-gram language model (LM) on the Arabic text with Kneser-Ney smoothing using KenLM \cite{kenlm}.
We tuned the models using pairwise ranking optimization (PRO) \cite{Hopkins:May:2011:PRO} with the length correction of \cite{nakov-guzman-vogel:2012:PAPERS}\footnote{For a broader discussion see also \cite{guzman-nakov-vogel:2015:CoNLL,nakov-guzman-vogel:2013:Short}.} with 1000-best lists, using the IWSLT'13 tuning dataset. 
We tested on the IWSLT'13 test dataset, which we decoded using the Moses~\cite{koehn2007moses} decoder to produce Arabic translations, which we then desegmented.

In order to ensure stability, we performed three reruns of MERT for each experiment,
and we report evaluation results averaged over the three reruns,
as suggested in \cite{Foster:2009}.

\subsection{Evaluation Setup}

One of the issues when translating into Arabic are the common mistakes in written Arabic text such as incorrect glyphs, punctuation, and other orthographic features of the language. 
%They are inconsistent and vary from one source to another depending on the author. 
With 29K pairs of different movies, it would be unrealistic to attempt to model them. Furthermore, the Arabic references provided by IWSLT in both the tuning and the testing sets are no exception from these random errors, which is an issue for scoring outputs of the systems when they are expected to produce proper Arabic. In fact, it is shown to affect the IWSLT baseline system score by more than two BLEU points with the NIST v13 scoring tool \cite{sajjadqcri}. Therefore, we normalize both the tuning and the testing references with the QCRI Arabic
Normalizer v3.0,\footnote{This is the official scoring method for the translation tracks into Arabic at IWSLT'13: \url{http://alt.qcri.org/tools/arabic-normalizer}} which use MADA's morphological analysis to output an enriched
%\footnote{We used the correct orthographic forms, as predicted by MADA's disambiguation, to ensure consistency.} 
form of Arabic; it also standardizes the digits to appear all in Arabic and converts all the punctuation to English punctuation to make it possible to tokenize. We apply the same normalization to each system's output after detokenization as a final post-processing step \cite{nakov2013parameter}.

\subsection{Experiments}

\subsubsection{Subtitle Alignment}

We produced two datasets with an overlap ratio threshold of 0.65, one with 1-1 alignments only (i.e., we only allow aligning one English segment to one Arabic segment) and the other one with 1-M alignments\footnote{We set $M$ to be at most 5 in order to prevent the algorithm from unreasonably iterating up to the last segment looking for a match.} (i.e., we allow aligning an English segment to a sequence of one or more consecutive Arabic segments). 

We choose the threshold of 0.65 roughly as a midpoint between the most strict condition, 1.0, which is not promising for the open nature of the data, or the less strict condition, 0.5, which creates the possibility of two adjacent segments being viable for a match with the same segment on the other side of the pair, and hence create duplications. Yet, this is only one case of presenting noise to the data, which will be contaminated with partial alignments, i.e., pairs with translations of only half of the other, when lowering the threshold.

We also produced a third 1-M alignment with a stricter (0.85) threshold that excluded pairs yielding less than 20\% matches between their original content. We assume that these pairs are likely to be out of sync. We also aim to test whether accounting for quality would enhance the translation performance even at the expense of quantity.

For comparison, we also trained an MT model on a bi-text produced by the best available subtitle alignment tool, \verb|subalign|, developed by Tiedemann. This tool includes more complex features such as auto-synchronization and matching segments at the sentence level~\cite{tiedemann2012opus}. It does not provide an English-Arabic dictionary for synchronization nor does it support the use of cognates for this language pair, so we set the parameters to default. 

The comparisons are in Table~\ref{table:compare}.
%\footnote{Note that the number of tokens in this table is calculated from the raw output of the preprocessing script, i.e., untokenized and unsegmented, and tokens are simply defined by surrounding white spaces while words are defined as unique tokens.}
The numbers show an 11.3\% increase in the total number of matched pairs and also a 5\% increase in the average sentence length in the 1-M model compared to the 1-1 model.
The strict 1-M model yields significant drop in the number of sentences and word tokens.
Finally, \verb|subalign| yields a bi-text that is about twice as large
as our bi-texts. Yet, as we will see below, this does not mean that it is of better quality.

\begin{table}
	\centering
	\begin{tabular}{lccc}
		{\bf Model} & {\bf \#Sentences}     & {\bf Avg. Len.} & \bf \#Tokens \\
		\hline
		1-1     & 13.3 M        & 6.3   & 83.8 M  \\
		1-M          & 14.8 M & 6.6     & 97.7 M  \\
		1-M Strict      & 11.1 M & 6.4 & 71.0 M \\ 
		\verb|subalign|   & 18.6 M & 9.6 & 178.6 M \\
		\hline
	\end{tabular}
	\caption{A comparison of the size of the produced corpora.}
	\label{table:compare}
\end{table}

%If there is room, I think you could say more about the alignment method here. I’m sure there is more detail to add. It would also be nice to see diagrams of a few alignments that illustrate features, problems etc. I don’t get much of a sense of how it actually works from your words.

\subsubsection{Machine Translation}

We tested the impact of adding various versions of our 
%movie subtitles 
bi-text to the training bi-text of the baseline system.
We used the movies bi-text in two different ways:
(a)~we built a separate phrase table, which then we interpolated with the phrase table of the baseline system \cite{nakov:2008:WMT,Nakov:2009:ISM,wang-nakov-ng:2012:EMNLP-CoNLL,Pidong:al:CL:2016}, and
(b)~we just concatenated our bi-text with the IWSLT'13 training bi-text in the baseline and trained on the concatenation.
We found that option (a) worked better on the 1-1 alignment model, and thus, we decided to use (a) for all models.

\begin{table}
	\centering
	\begin{tabular}{llc}
		{\bf System} & {\bf Data}     & {\bf BLEU}  \\ \hline
		Baseline     & IWSLT        & 11.90       \\
		\hline
		1-1          & IWSLT+SUB & 13.59       \\
		1-M          & IWSLT+SUB & {\bf 13.98} \\
		1-M Strict      & IWSLT+SUB & 12.89       \\ 
		\hline
		\verb|subalign|    & IWSLT+SUB & 13.69 \\ \hline
	\end{tabular}
	\caption{Evaluation results for various alignment models.}
		\label{table2}
\end{table}

Table~\ref{table2} shows a comparison between the baseline system and the three proposed
systems that combine the IWSLT data from the baseline with the subtitles corpora using
the corresponding alignment models. The baseline system scored 11.9 BLEU points, which is a replicate of the same results reported on the IWSLT'13 official page.\footnote{\url{https://wit3.fbk.eu}} 

We can see in the table that the simplest model's system, which used the 1-1 alignment, had an advantage over the baseline by only 0.61 BLEU points absolute when we concatenated the texts of both the baseline and the subtitles for training the language model. 

We ran another experiment with the same parameters except that in this run, we created separate LMs for the two corpora and then we interpolated them to
optimize their weights. This improved the score to 1.71 BLEU points above the baseline
score. Furthermore, the 1-M model's system yielded an even better score, 13.98 BLEU points, and is the best system overall. On the other hand, the strict version of the
1-M model only yielded 12.89 BLEU points, which is better than the baseline but worse than the simplest approach, the 1-1 model.

On the other hand, the 1-M strict model's SMT output yielded a better result than the 1-1 model, 12.89 BLEU points, but it is still lower than the original 1-M model. This may imply that SMT independently filters out noise because of its statistical nature and, yet, the translation model may make use of the excess data to compute more accurate statistics.

Finally, we tried to produce a corpus using the OPUS Uplug \verb!subalign! tool to align the data. The tool comes with its own preprocessor, which we had to use in order to produce output in XML format as expected by the alignment script. Although this tool allows synchronization of subtitles, it requires a dictionary and/or a language model for this feature to work, which were not provided for Arabic and the time was not feasible to prepare any. After all, it would be more relevant to compare the output of that tool using only features matching our own model's. \verb!subalign! managed to harvest about 18M parallel segment lines with longer average sentence length (English) from our subtitle collection, although for Arabic it was contaminated with gibberish caused by poor handling of the UTF-8 character encoding. Several, workarounds were attempted by force re-encode the text, fixing parts of the source code relevant to encoding and enforcing the encoding flag to UTF-8, but these were ineffective. This is also despite the fact the we applied the \verb!dos2unix! utility to the raw text as instructed to eliminate any special DOS-related characters. Thus, the BLEU evaluation for the SMT pipeline using this corpus is not comparable to the rest.

\section{Conclusion and Future Work}
\label{sec:discuss}

%Tabel \ref{} shows that the addition of the subtitle data improves the translation performance by at least 1.69 BLUE over what baseline?. The 1-M \emph{Strict} model's output returned a lower score than the 1-M's. This may imply that SMT anyhow filters out noise because of its statistical nature and, yet, the TM what’s that? may make use of the excess data to compute more accurate probabilities.

%ur model also outperforms \verb|subalign| by 0.29 BLUE despite the significant advantage of quantity ?? “significant extra material used”?? by the latter. This could be due to the complications caused by concatenating segments to form longer sentences despite the inconsistencies of the subtitles' sentence boundary punctuations. Alternatively, it could be due to different subtitling traditions. This suggests that a simpler approach is more beneficial for translation quality.

%Needs more detailed analysis of performance than this. 

%We have presented a low-cost approach for aligning a substantial amount of overlooked multilingual data with the potential to improving spoken SMT, as proven earlier. Our technique uses a time overlap alignment approach on English-Arabic SMT, and we find that quantity is a crucial factor for this translation direction at least, given the scarcity of parallel data.

We have presented our efforts towards getting better resources for English-Arabic machine translation of spoken text. We created the largest English-Arabic bi-text to date, and we developed a subtitle segment alignment algorithm, which outperformed the best rivaling tool, \verb|subalign|. The evaluation results have shown that our dataset, combined with our alignment algorithm, yielded an improvement of over two BLEU points absolute over a strong spoken SMT system.

In future work, we would like to incorporate more sources of information in the process of alignment, e.g., cognates, translation dictionaries, punctuation, numbers, dates, etc. 
%We also wonder whether we could make use of metadata about a movie, e.g., from IMDB.
Adding some language-specific information might be interesting too, e.g., linguistic knowledge.
%, or general statistics about how many letters a typical segment accepts in a language.
We also plan to 
%make our segment alignment tool more intelligent, e.g., by teaching it 
teach our tool
to recognize advertisements, which are typically placed at the beginning or at the end of the file. Another interesting research direction is choosing the best pair of subtitles, in case multiple versions of subtitles exist for the same movie for the source and/or for the target language.

\bibliography{main.bib}
\bibliographystyle{plain}

\end{document}